\documentclass[letterpaper]{article}
\usepackage[nonatbib]{nips_2017}
\usepackage{times}
\usepackage{helvet}
\usepackage{courier}
\usepackage{url}
\usepackage{graphicx}
\usepackage{hyperref}

\usepackage{amsmath}
\usepackage{amssymb}
\begin{document}
\title{A Classifying Variational Autoencoder with Application to Polyphonic Music Generation}
\author{Jay A. Hennig, Akash Umakantha, Ryan C. Williamson\\Center for the Neural Basis of Cognition, Machine Learning Department\\Carnegie Mellon University, Pittsburgh, PA, USA\\ \texttt{\{jhennig, aumakant, rcw1\}@andrew.cmu.edu}}
\maketitle
\begin{abstract}
The variational autoencoder (VAE) is a popular probabilistic generative model. However, one shortcoming of VAEs is that the latent variables cannot be discrete, which makes it difficult to generate data from different modes of a distribution. Here, we propose an extension of the VAE framework that incorporates a classifier to infer the discrete class of the modeled data. To model sequential data, we can combine our Classifying VAE with a recurrent neural network such as an LSTM. We apply this model to algorithmic music generation, where our model learns to generate musical sequences in different keys. Most previous work in this area avoids modeling key by transposing data into only one or two keys, as opposed to the 10+ different keys in the original music. We show that our Classifying VAE and Classifying VAE+LSTM models outperform the corresponding non-classifying models in generating musical samples that stay in key. This benefit is especially apparent when trained on untransposed music data in the original keys.
\end{abstract}

\section{Introduction}

For decades, researchers have approached the task of algorithmic music generation using computational models \cite{papadopoulos1999ai}. One common approach to this task is to generate samples sequentially using a probability model, using a corpus of training data to learn a probability distribution of the most likely notes to be played at each time step. This approach has been used successfully to compose music in the style of Bach or other composers \cite{liang2016bachbot}. However, most previous work using neural networks has avoided modeling the key of the music, usually by transposing all songs in the corpus into only one or two keys \cite{mozer1994neural,paiement2009probabilistic,rnn-rbm,vohra2015modeling,liang2016bachbot,johnson2017generating}. Alternatively, instead of transposing songs into only one or two keys, one could augment the training data by copying each song into multiple keys. However, both approaches typically have one or many of the following shortcomings. First, these models may not allow for the user to explicitly control which key the generated samples are in (e.g., C major versus C minor). Second, as a result of the previous point, the samples generated by these models may alternate between different keys as the generating process continues, resulting in unappealing music samples. Finally, music likely has different probabilistic structure in different keys, owing to various technical or subjective considerations of the composer (e.g., the use of different musical temperaments, or the idea that a song in C major ``feels'' different than a song written in Db major; see \cite{young1991key}). To resolve these difficulties, we introduce an approach that allows for control over the key of generated musical samples and does not require first transposing the training data into a limited number of keys.

Many recent approaches at generating polyphonic music have used variational autoencoders (VAEs) \cite{kingma2013,rezende2014,fabius2014variational,hadjeres2017glsr}, a popular deep generative model that learns both a recognition and generation model for the observed data. However, though VAEs can successfully generate various forms of data such as digits and faces, when the data is multi-modal, VAEs do not provide an explicit mechanism for specifying which mode or class a generated sample should come from. For example, in the case of digit generation, one may wish to ensure the model generates a `2' and not a `4'. Analogously, in the case of music generation, one may wish to ensure the model generates music consistent with the key of C major and not C minor. One option is to provide the class label as an additional input to the model, as in a conditional VAE \cite{sohn2015learning,walker2016uncertain}. However, in the case of music generation, it may be desirable for the model to be able to infer the class label using the sequence history (e.g., when improvising with a human musician). Because VAEs cannot perform inference over discrete-valued latent variables, one common approach is to model the latent distribution as a mixture distribution such as a mixture of Gaussians \cite{vaetutorial}. However, this approach is difficult to train and often not successful in practice \cite{advAuto}.

Our model, which we call a Classifying VAE \footnote{Code and examples of generated music are available at \url{https://mobeets.github.io/classifying-vae-lstm/}}, extends the standard VAE by including a classifier to infer the class of each data point. This classifier is trained concurrently with the recognition and generation models of the standard VAE. By modeling the class probabilities as a Logistic Normal distribution, we can use the reparameterization trick of Stochastic Gradient Variational Bayes to sample from this class distribution during training \cite{kingma2013,rezende2014}. This enables efficient inference while still allowing for control over the class of generated samples. The Classifying VAE can be further extended to model the temporal structure of music by including recurrent neural networks such as long short-term memory (LSTM) networks \cite{hochreiter1997long} in the recognition and generative models, following previous work \cite{STORN,latent_rNN,SRNN}. When applied to music generation, we show that our model can be trained without first transposing the training data into different keys, and allows for generating music that stays in a user-specified key.

This paper is structured as follows. First, we introduce the task of modeling polyphonic music, and provide background on previous approaches using VAEs with LSTMs. Next, we introduce the Classifying VAE and Classifying VAE+LSTM, which allow us to generate data from one of several discrete classes. We then apply these models to music generation and show that the Classifying VAE learns an interpretable latent space. Finally, we show that the Classifying VAE and Classifying VAE+LSTM generate musical sequences that stay in key more often than similar models when trained on songs in their original keys.

\section{Preliminaries}

\subsection*{Polyphonic music}

\begin{figure}[ht]
\centering
\includegraphics[width=0.49\textwidth]{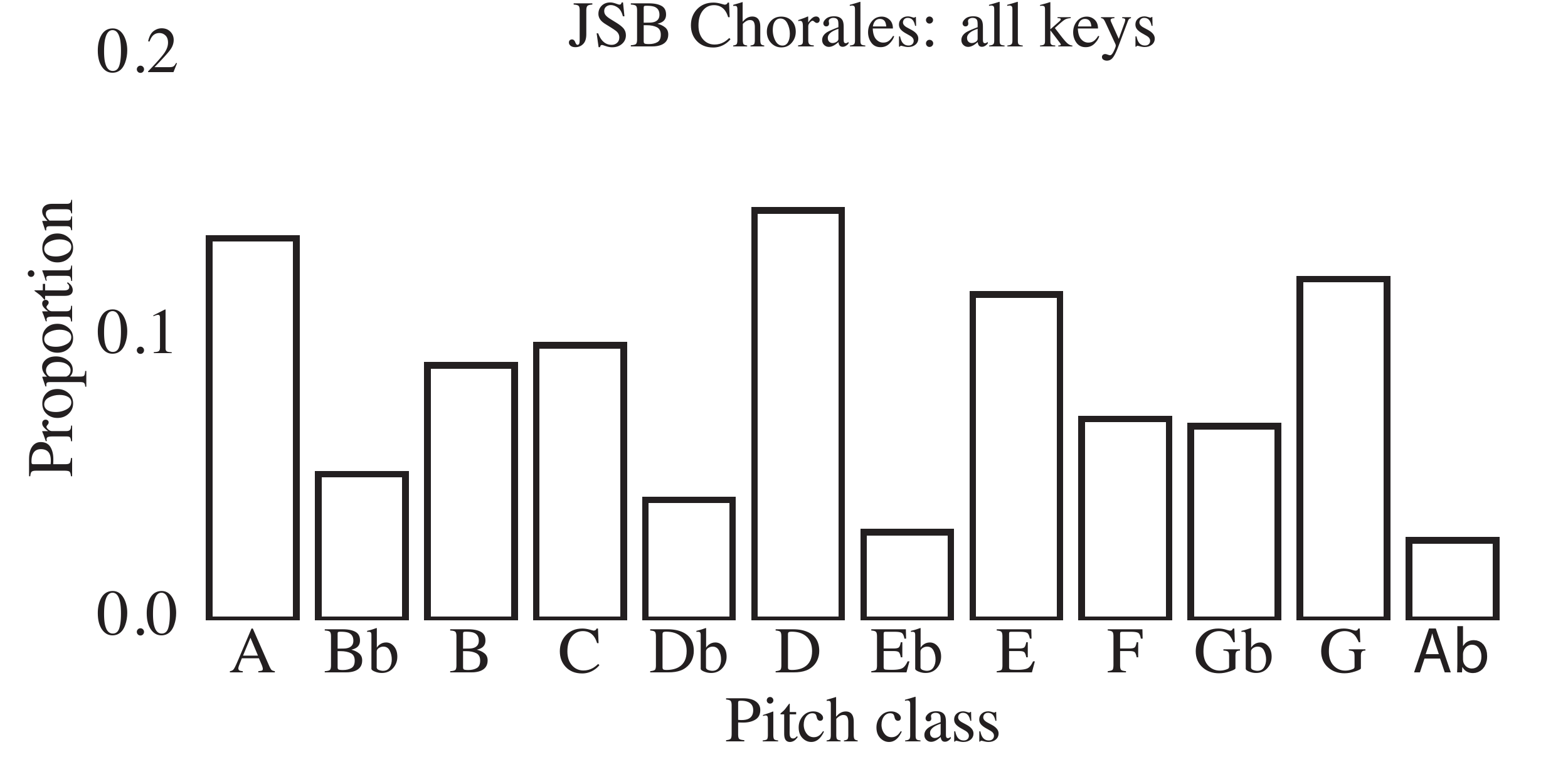}
\includegraphics[width=0.49\textwidth]{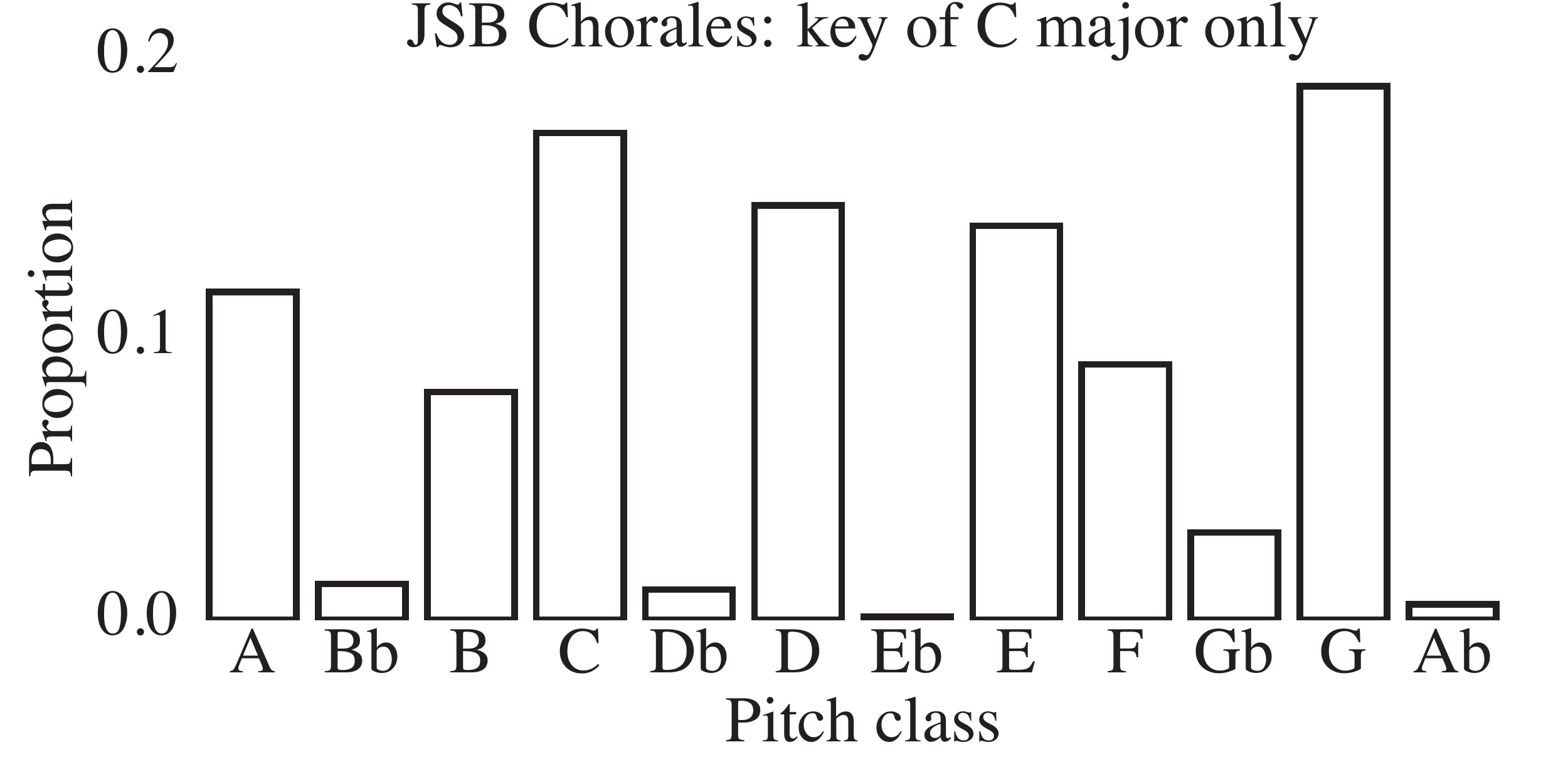}
\caption{The distribution of pitch classes in a music corpus depends on the keys of the songs considered. Left: Distribution of pitch classes present in all songs in JSB Chorales. Right: Distribution of pitch classes present in all songs in the key of C major in JSB Chorales.}
\label{fig:pitches}
\end{figure}

Western music uses a set of 12 pitch classes $(A, Bb, B, C, Db, D, Eb, E, F, Gb, G, Ab)$ (Figure \ref{fig:pitches}, top). Most songs are written using only a subset of these classes depending on what key the song is written in. The key of a song refers to the central tonic note (e.g., $C$, $D$, etc.), along with a mode (e.g., major, minor, etc.). For example, music written in the key of C major will tend to use all notes without sharps or flats $(A, B, C, D, E, F, G)$ (Figure \ref{fig:pitches}, bottom), with ``C'' being the tonic note that most melodies resolve on. Music in the key of C minor, by contrast, will have the same tonic note (C) but a different overall set of notes $(C, D, Eb, F, G, Ab, Bb)$. Overall, there are 24 possible keys, corresponding to the twelve tonic notes in either the major or minor mode. However, each major key has a relative minor key that uses the same pitch classes. For example, the relative minor key of C major is A minor $(A, B, C, D, E, F, G)$. For simplification, we treat these pairs of major and relative minor keys as the same key class, resulting in 12 distinct key classes.

In this paper we model music data as a series of 88-dimensional binary vectors, $X_t \in \{0,1\}^{88}$, where the $j^{th}$ entry, $X_t^j$, can be thought of as representing whether the $j^{th}$ key on an 88-key piano was played at time $t$. In polyphonic music, multiple notes may be played simultaneously (i.e., $\sum_{j=1}^{88} X_t^j \in \{0, 1, ..., 88\}$). We define $X = \{ X_t \mid t = 1, ..., T\}$ as a music sequence with length $T$. As an example of how key determines which notes are likely to be played, let $w \in \{0,1\}$ be the key of $X$, and suppose that $w = 0$ refers to the key of C major. Then for any $t \in \{1, ..., T\}$ and $j \in \{1, ..., 88\}$, we know that $P(X^j_t \mid w = 0) \approx 0$ if the pitch class of note $j$ is not in $\{A, B, C, D, E, F, G\}$. While songs do occasionally change key, we assume that the key is constant within short segments of music, e.g., when $T$ is sufficiently small.

\subsection*{The variational autoencoder}
Variational autoencoders \cite{kingma2013,rezende2014} provide a flexible framework for generating samples from complex distributions using the Stochastic Gradient Variational Bayes (SGVB) estimator. The VAE models observed data, $X$, as a nonlinear transformation of unobserved latent variables, $z$. A recognition (encoding) network is trained to infer the posterior distribution of likely latent variables given the observed data, while a generating (decoding) model is trained to transform samples from this posterior distribution to match the observed data. The joint distribution under consideration is the following:
\begin{align*}
p_\theta(X, z) = p_\theta(X \mid z)p_\theta(z)
\end{align*}
where $\theta$ are a set of generative parameters. The prior on the latent variables, $p_\theta(z)$, typically takes a simple form (e.g., a standard multivariate Gaussian distribution) to allow for straightforward model fitting and data generation. The VAE model is trained to learn the generating model, $p_\theta(X\mid z)$, as well as a variational approximation of the recognition model, $p_\theta(z \mid X)$. The main idea is to first write the following equality involving the total marginal likelihood, $p_\theta(X) = \int p_\theta(X \mid z)p_\theta(z)\partial z$:
% The main idea is to first rewrite the KL divergence ($\mathcal{D}$) between $p_\theta(X \mid z)$ and the total marginal likelihood, $p_\theta(X) = \int p_\theta(X \mid z)p_\theta(z)\partial z$, as the following:
% \begin{align*}
% \mathcal{D}[ q_\phi(z) \| p_\theta(z \mid X)] = \mathbb{E}_{z \sim q}[ \log q_\phi(z) - \log p_\theta(z \mid X)]
% \end{align*}
% where $q$ is an arbitrary distribution. After rearranging:
\begin{align*}
\log p_\theta(X) - &\mathcal{D}[ q_\phi(z \mid X) \| p_\theta(z \mid X)] = \mathbb{E}_{z \sim q_\phi(z \mid X)}[\log p_\theta(X \mid z)] - \mathcal{D}[ q_\phi(z \mid X) \| p_\theta(z)]
\end{align*}
where $\mathcal{D}$ is the KL divergence, $q_\phi(z\mid X)$ is our variational approximation to the posterior with parameters $\phi$, and the right-hand side is called the evidence lower-bound, or ELBO. To maximize the marginal likelihood, $p_\theta(X)$, we can maximize the ELBO, provided we ensure that $\mathcal{D}[ q_\phi(z \mid X) \| p_\theta(z \mid X)]$ is small. The standard approach, which we use here, is to assume that $q_\phi(z \mid X)$ is a multivariate Gaussian with mean and diagonal variance computed by a multilayer perceptron given the input $X$. We can compute $\mathbb{E}_{z \sim q_\phi(z \mid X)}[\log p_\theta(X \mid z)]$ using a reparameterization trick, which allows the ELBO to be maximized using stochastic gradient descent on $\theta$ and $\phi$ with gradients computed via backpropagation. This reparameterization consists of rewriting $z \sim q_\phi(z \mid X)$ as $z = g_\phi(X, \epsilon)$, where $g_\phi$ is some continuous function with respect to $X$, and $\epsilon$ is a sample from a random (untrained) noise distribution. In the case of $q_\phi(z \mid X)$ above with mean $\mu_\phi(X)$ and diagonal covariance $\sigma_\phi^2(X)$, we can write $z = \mu_\phi(X) + \sigma_\phi(X)\epsilon$, where $\epsilon \sim \mathcal{N}(0, I)$. However, one drawback of this approach is that it does not apply for discrete latents. This is problematic if we want to generate music samples from a discrete latent key.

\section{The Classifying Variational Autoencoder}

To allow the VAE to infer the class of its generated data, we incorporate an additional continuous latent variable, $w$, that represents the inferred probability of the data belonging to each of $d$ distinct classes (e.g., $d$ is the number of keys). The joint distribution we consider is
\begin{align*}
p_\theta(X, z, w) = p_\theta(X \mid z, w)p_\theta(z)p_\theta(w)
\end{align*}
where $X$ is the observed data, $z$ and $w$ are unobserved latent variables, and $z$ and $w$ assumed to be independent.

Above, $w$ is a vector of dimension $d$, the number of classes. We assume that during training we are given $X$ as well as $w$, while at test time (e.g., for generation) we have only $X$. Because a variational autoencoder cannot infer discrete latent variables, our approach is to treat $w$ as a set of class probabilities rather than as a discrete categorical variable. Our goal during training is to learn the full posterior of class probabilities, $p_\theta(w \mid X)$, by training a classifier network to match both the mean and variance of this posterior. Samples from this inferred posterior are then fed into the recognition model, potentially providing an extra source of stochasticity.

We now follow the main idea of the variational autoencoder (as above). First, we construct a variational lower-bound on the log of the marginal likelihood $p_\theta(X) = \int p_\theta(X \mid z, w)p_\theta(z)p_\theta(w) \partial w \partial z$. We can write the following equality:
\begin{align*}
\log p_\theta(X) &- \mathcal{D}[ q_\phi(z, w \mid X) \| p_\theta(z, w \mid X)] \\ &= \mathbb{E}_{(z, w) \sim q_\phi(z,w \mid X)}[\log p_\theta(X \mid z, w)] - \mathcal{D}[ q_\phi(z, w \mid X) \| p_\theta(z, w)]
\end{align*}
Applying the chain rule on the KL terms, we can rewrite the left-hand side as:
\begin{align*}
\log p_\theta(X) & - \mathbb{E}_{w \sim q_\phi(w \mid X)}[\mathcal{D}[ q_\phi(z \mid X, w) \| p_\theta(z \mid X, w)]] - \mathcal{D}[ q_\phi(w \mid X) \| p_\theta(w \mid X)]
\end{align*}
We can similarly rewrite the right-hand side:
\begin{align*}
\mathbb{E}_{(z,w) \sim q_\phi(z,w \mid X)}&[\log p_\theta(X \mid z, w)] \\&-\mathbb{E}_{w \sim q_\phi(w \mid X)}[\mathcal{D}[ q_\phi(z \mid X, w) \| p_\theta(z)]] - \mathcal{D}[ q_\phi(w \mid X) \| p_\theta(w)] = -\mathcal{L}_{VAE}(X)
\end{align*}

where we have also applied the independence of $w$ and $z$. As in a standard VAE, $p_\theta(z)$ and $p_\theta(w)$ are priors, while $q_\phi(w \mid X)$ and $q_\phi(z \mid w, X)$ are variational approximations to the true posteriors. We aim to maximize the marginal likelihood, $\log p_\theta(X)$, by maximizing the right-hand side using stochastic gradient descent on $\theta$ and $\phi$.

To keep our lower-bound on the marginal likelihood tight, we must also ensure the KL divergences on the left-hand side are small. As in the standard VAE, here we suppose we do not have access to the true posterior $p_\theta(z \mid X, w)$. However, we suppose that when training we \textit{do} have access to the true class (a delta function), which we denote $\tilde{w}$. Thus, instead of minimizing $\mathcal{D}[ q_\phi(w \mid X) \| p_\theta(w \mid X)]$, we can minimize the categorical cross-entropy loss between $q_\phi(w \mid X)$ and $\tilde{w}$, encouraging $q_\phi(w \mid X)$ to classify $X$. For this reason, we call our model a Classifying VAE. The final objective ($\mathcal{L}$) is then:
\begin{align*}
\mathcal{L}(X) = \mathcal{L}_{VAE}(X) + \alpha \mathbb{E}_{w \sim q_\phi(w \mid X)}[\mathcal{L}_c(w; \tilde{w})]
\end{align*}
where in the final term, $\mathcal{L}_c(w; \tilde{w})$ is the categorical cross-entropy loss between a sampled $w \sim q_\phi(w \mid X)$ and the true class $\tilde{w}$, while $\alpha$ is a hyperparameter. Note that when $\alpha = 0$, this is the objective for a standard VAE.

The key idea of our model is that samples from $q_\phi(w \mid X)$ are used in the recognition model for $z$ while also affecting the classification loss. The goal is that improved classification of $X$ will lead to better reconstruction of $X$ as well as control over the class of generated samples. As with the standard VAE, we use the Stochastic Gradient Variational Bayes (SGVB) estimator of the ELBO, whereby we draw samples of $w \sim q_\phi(w \mid X)$ and $z \sim q_\phi(z \mid w, X)$ for each minibatch during training. One of the limitations of the VAE is that the latent variables $z$ and $w$ cannot be discrete. We must be able to apply the ``reparameterization trick'' and generate samples of $z$ and $w$ as differentiable, deterministic functions of $X$ and some auxiliary variables $\epsilon$ with independent marginals. We can choose $q_\phi(z \mid w, X)$ to be a standard multivariate Gaussian as we did with the standard VAE. However, $q_\phi(w \mid X)$ is intended to be the inferred posterior probabilities that $X$ belongs to each class, but the natural choice of $q_\phi(w \mid X)$ as Dirichlet is not compatible with the reparameterization trick. Instead, we parameterize $q_\phi(w \mid X)$ as a Logistic Normal distribution \cite{atchison1980logistic} with mean and diagonal covariance computed by the output of a multilayer perceptron given the input $X$. Samples $w \sim q_\phi(w \mid X) = \mathcal{L}\mathcal{N}(\mu_\phi(X), \sigma_\phi^2(X))$ have the property that $0 \leq w_i \leq 1$ for $i = 1, ..., d$, and $\sum_{i=1}^d w_i = 1$, so that $w$ can be interpreted as the estimated probabilities that $X$ belongs to each of the $d$ classes. Critically, samples from the Logistic Normal distribution can be generated deterministically from samples drawn from a Normal distribution with the same parameters. Specifically, if $y \sim \mathcal{N}(\mu, \Sigma)$ is a sample from a Normal distribution with $y \in \mathbb{R}^{d-1}$, then $w \sim \mathcal{L}\mathcal{N}(\mu, \Sigma)$ is a sample from a Logistic Normal if we set $w_j = \frac{e^{y_j}}{1 + \sum_{j=1}^{d-1} e^{y_j}}$ for $j = 1, ..., d-1$ and $w_d = \frac{1}{1 + \sum_{j=1}^{d-1} e^{y_j}}$. This allows us to sample both $w \sim q_\phi(w \mid X)$ and $z \sim q_\phi(z \mid X,w)$ using the reparameterization trick, enabling us to train the Classifying VAE using SGVB.
\begin{figure}[ht]
\centering
\begin{minipage}{0.2\textwidth}
 \centering
 \textbf{Inference model for $w$}\\
\vspace{5mm}\includegraphics[width=0.3\textwidth]{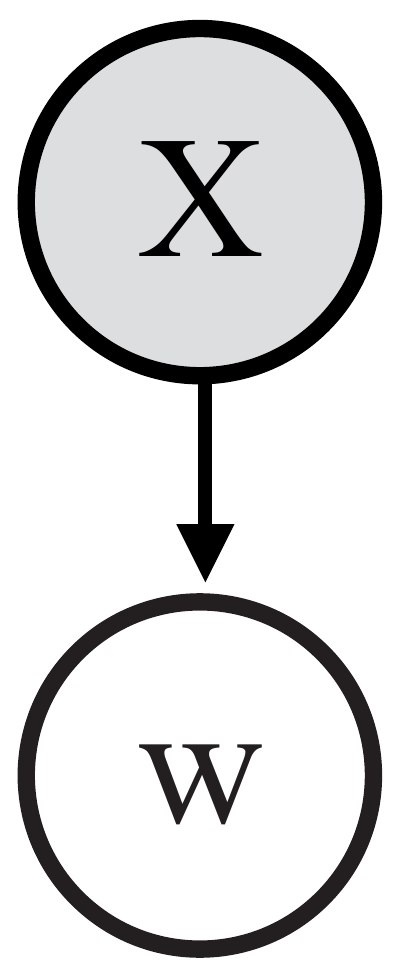}
\end{minipage}\qquad
\begin{minipage}{0.2\textwidth}
 \centering
 \textbf{Inference model for $z_t$}\\
\vspace{3mm}\includegraphics[width=0.8\textwidth]{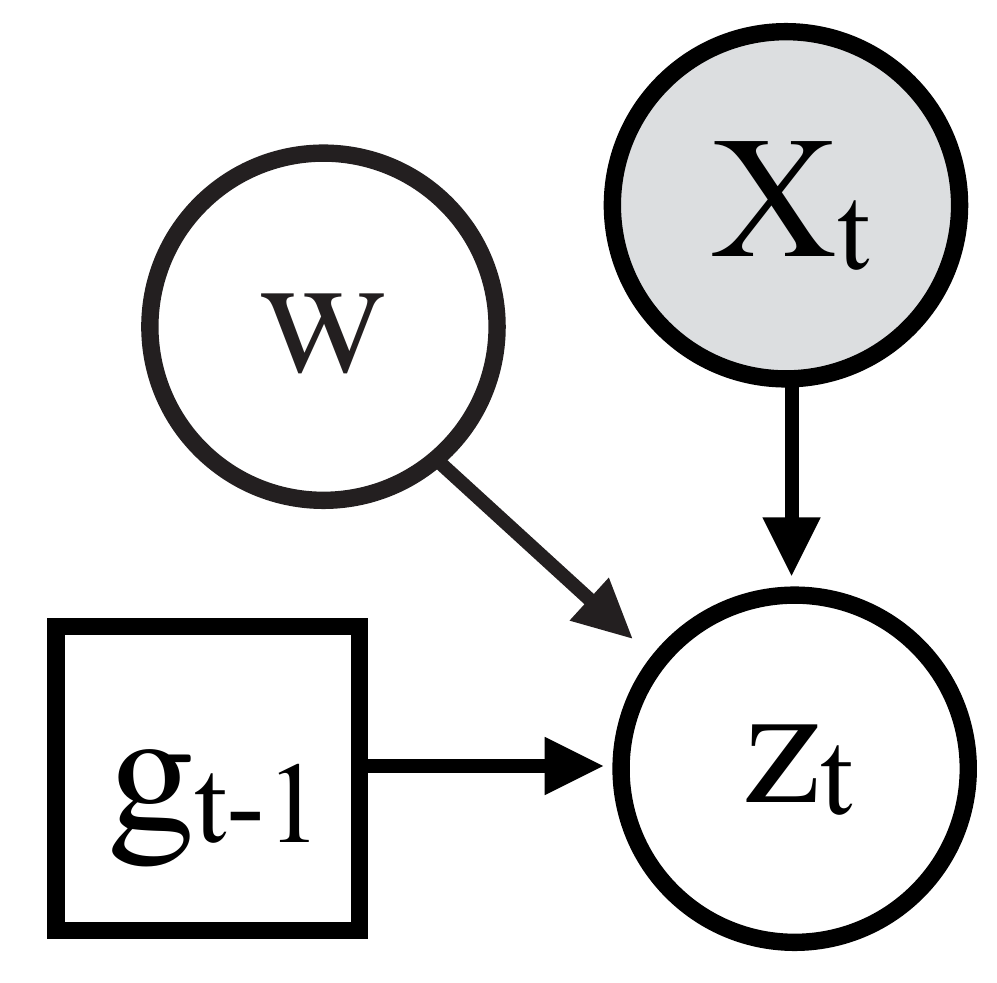}
\end{minipage}\hspace{1cm}
\begin{minipage}{0.2\textwidth}
 \centering
 \textbf{Generative model for $X_t$}\\
\vspace{4mm}\includegraphics[width=1.1\textwidth]{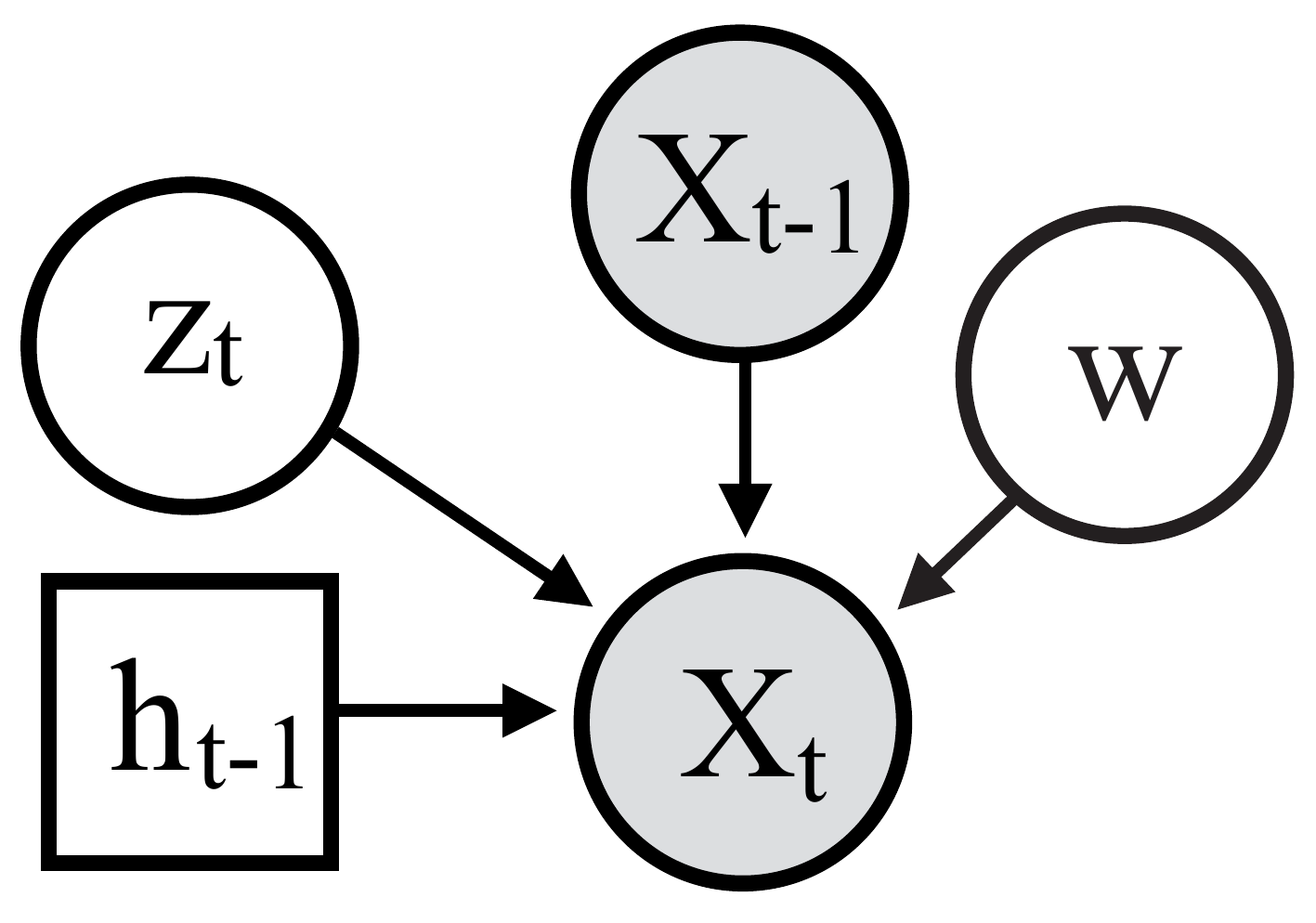}
\end{minipage}
\caption{From left to right, the inference model for $w$, the inference model for $z_t$, and the generative model for $X_t$ in the Classifying VAE+LSTM. $w$ is a continuous latent variable representing the probability that the sequence $X$ came from one of $d$ discrete classes. $z_t$ is a continuous latent variable, inferred using an encoder LSTM with deterministic hidden state $g$. $X_t$ is an observed discrete variable, and is generated using the decoder LSTM with deterministic hidden state $h$.}
\label{fig:models}
\end{figure}
\subsection{Modeling sequences with a Classifying VAE+LSTM}
The VAE (and Classifying VAE) attempts to ``autoencode'' or reconstruct an input $X$ using a recognition and generating network. When $X$ is a sequence, one may still use a standard VAE by autoencoding each time step $X_t$ independently.
% The additions of the Classifying VAE+LSTM to the generative model of STORN are depicted in blue in the right panel of Figure \ref{fig:models}.
% Figure \ref{fig:models} (right panel, in black) shows a diagram of the STORN generative model.
Recent extensions of VAEs to sequence data incorporate recurrent neural networks (RNN) such as long short-term memory (LSTM) networks into both the encoder and decoder networks of a VAE \cite{STORN,latent_rNN,SRNN}. Broadly, this allows the encoding and decoding networks to be conditioned on the sequence history at each time step. We refer to these models as recurrent variational autoencoders, or VAE+LSTM. Here for simplicity we focus on the STORN model \cite{STORN}, where the sampled latent variables $z_t$ are assumed to be sampled independently over time from a stationary prior. In our experiments below we extend STORN to a Classifying VAE+LSTM analogously to how we extend the VAE to a Classifying VAE.
\section{Methods}
\subsection{Data}
We applied our method to generate sequences of polyphonic music in the form of piano-roll data. For training data we use both the entire corpus of 382 four-part harmonized chorales by J.S. Bach (``JSB Chorales''), and a collection of classical music by a variety of composers (``Piano-midi.de''), both obtained from the authors of \cite{rnn-rbm}. Following previous work \cite{rnn-rbm}, this corpus was discretized and converted to piano-roll notation, so that at each time step of the music, $X \in \{0,1\}^{88}$, is a binary vector denoting which of 88 notes (from A0 to C8) was played at that time.

One of the most popular algorithms for detecting the key of a musical sequence is the Krumhansl-Schmuckler algorithm \cite{krumhansl}. This algorithm finds the proportion of each pitch class present in a sequence, and compares this with the proportions expected in each key. We used the implementation of this algorithm in the Python package music21 \cite{music21} to establish the ground-truth key of each musical sequence in our training data. For the purposes of our experiments, we treated pairs of major and relative minor keys as the same key class (e.g., C major and A minor). For comparison with previous work on these data sets, we also present results for models trained on each corpus where every song in a major key was transposed to C major and every song in a minor key was transposed to C minor \cite{rnn-rbm,STORN,vohra2015modeling,lyu2015modelling,johnson2017generating}.

\subsection{Implementation}

\textit{Architecture.} For the Classifying VAE, we assume the following priors on $w$ and $z_t$ for $t = 1, ..., T$:
\begin{align*}
p_\theta(w) &= \mathcal{L}\mathcal{N}(0, I),\quad p_\theta(z_t) = \mathcal{N}(0, I)
\end{align*}
For our posterior approximations, we use:
\begin{align*}
q_\phi(w \mid X) &= \mathcal{L}\mathcal{N}(\mu_{w,\phi}(X), \mbox{diag}[\sigma_{w, \phi}^2(X)])\\
q_\phi(z_t \mid X_t, w) &= \mathcal{N}(\mu_{z,\phi}(X_t, w), \mbox{diag}[\sigma^2_{z,\phi}(X_t, w)])
\end{align*}
where $\mu_{w,\phi}$ and $\sigma^2_{w,\phi}$ are implemented as the outputs of a neural network (the ``classifier''), and $\mu_{z,\phi}$ and $\sigma^2_{z,\phi}$ are implemented as the outputs of a different neural network (the ``encoder''). Because the outputs of the encoder are a function of both $X$ and $w$, during training we draw a single sample $w \sim q_\phi(w \mid X)$ as described previously in order to compute $\mu_{z,\phi}(X_t, w)$ and $\sigma^2_{z,\phi}(X_t, w)$ for $t = 1, ..., T$. Finally, following \cite{STORN} we assume for $i = 1, ..., 88$:
\begin{align*}
p_\theta(X_t^{i} \mid w, z_t, X_{t-1}) = \mbox{Bernoulli}(\pi_{\phi}^{i}(w, z_t, X_{t-1}))
\end{align*}
where the probabilities $\pi_\phi^{i}$ are the output of a third neural network (the ``decoder'') given a single sample of $w$ and $z_t$ from the above $q_\phi$ distributions, as well as the previous time step $X_{t-1}$. For the classifier, encoder, and decoder networks, we use multilayer perceptrons (one for each network) with one hidden layer and ReLu activation functions. The outputs of the decoder network are passed through a sigmoid nonlinearity to constrain the values between $0$ and $1$.

We also implement a standard VAE \cite{kingma2013,rezende2014}, equivalent to ignoring $w$ in all equations above. For our standard VAE+LSTM, we implement a network similar to STORN \cite{STORN}. In this case, the encoder and decoder networks are each replaced with an LSTM followed by a single dense layer, so that the equations above are also conditioned on the hidden states of the LSTMs. The Classifying VAE+LSTM is similar except for the addition of a classifier, implemented analogously to the Classifying VAE (see Figure \ref{fig:models}). Note that the classifier is given an entire sequence $X$ of length $T$ to classify the key. For the Classifying VAE, $T=1$, while for the Classifying VAE+LSTM, we treat the sequence length as a hyperparameter.

\textit{Training.} All models were implemented using Keras with a Tensorflow backend \cite{chollet2015keras,tensorflow2015-whitepaper}. Weights were initialized to a random sample $\mathcal{N}(0, 0.01)$, and trained with stochastic gradient descent using Adam \cite{adamOpt} ($\alpha = 0.001$, $\beta_1 = 0.9$, $\beta_2 = 0.999$, $\epsilon = 1 \times 10^{-8}$). We found using weight normalization \cite{weightNorm} resulted in faster convergence. As suggested in other work \cite{bowman2015generating}, we also gradually introduced the KL terms into the loss function during training.

\begin{figure}[ht]
 \centering
 \begin{minipage}{0.29\textwidth}
 \centering
 \textbf{VAE}\\
\includegraphics[width=\textwidth]{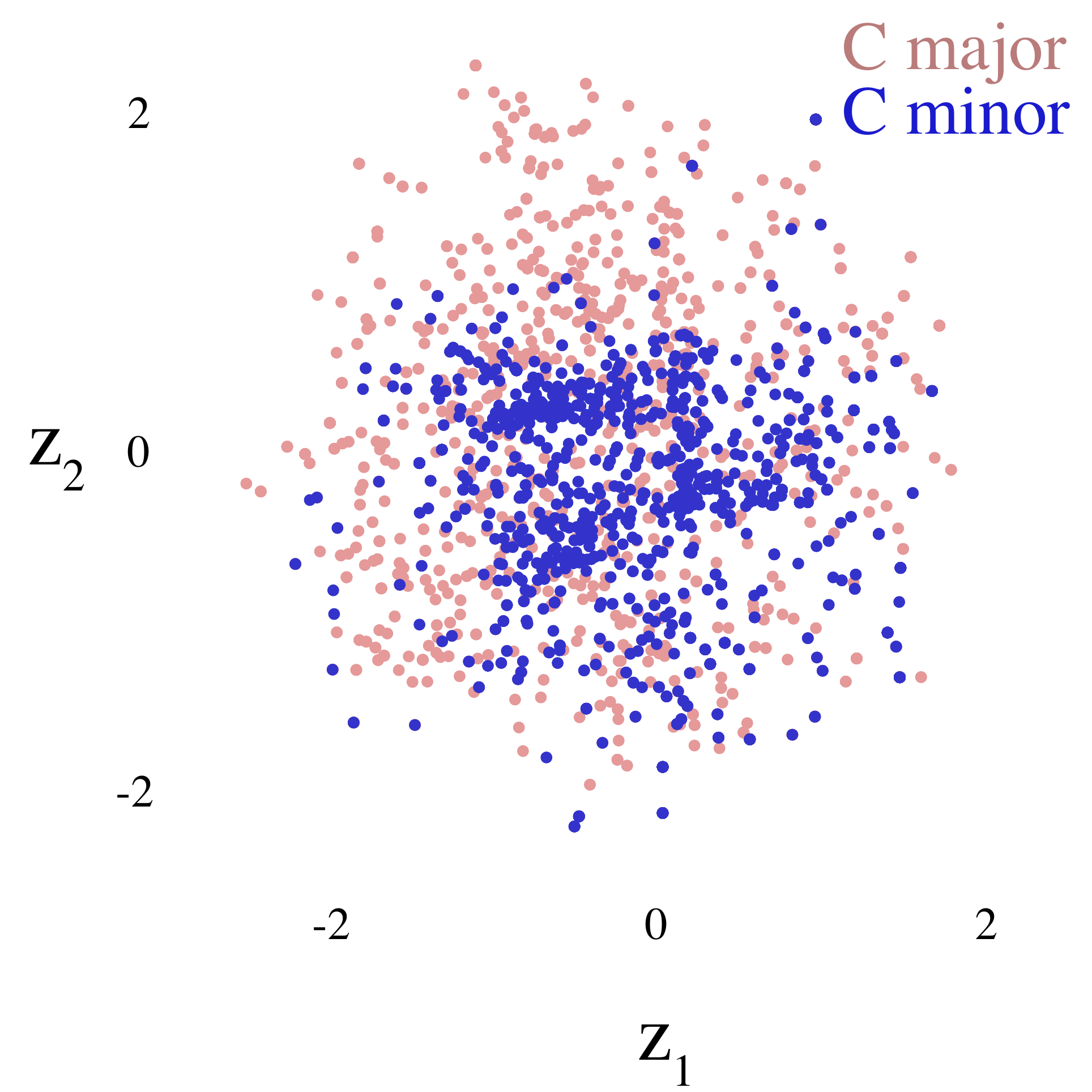}
\end{minipage}\qquad
 \begin{minipage}{0.29\textwidth}
 \centering
 \textbf{Classifying VAE}\\
\includegraphics[width=\textwidth]{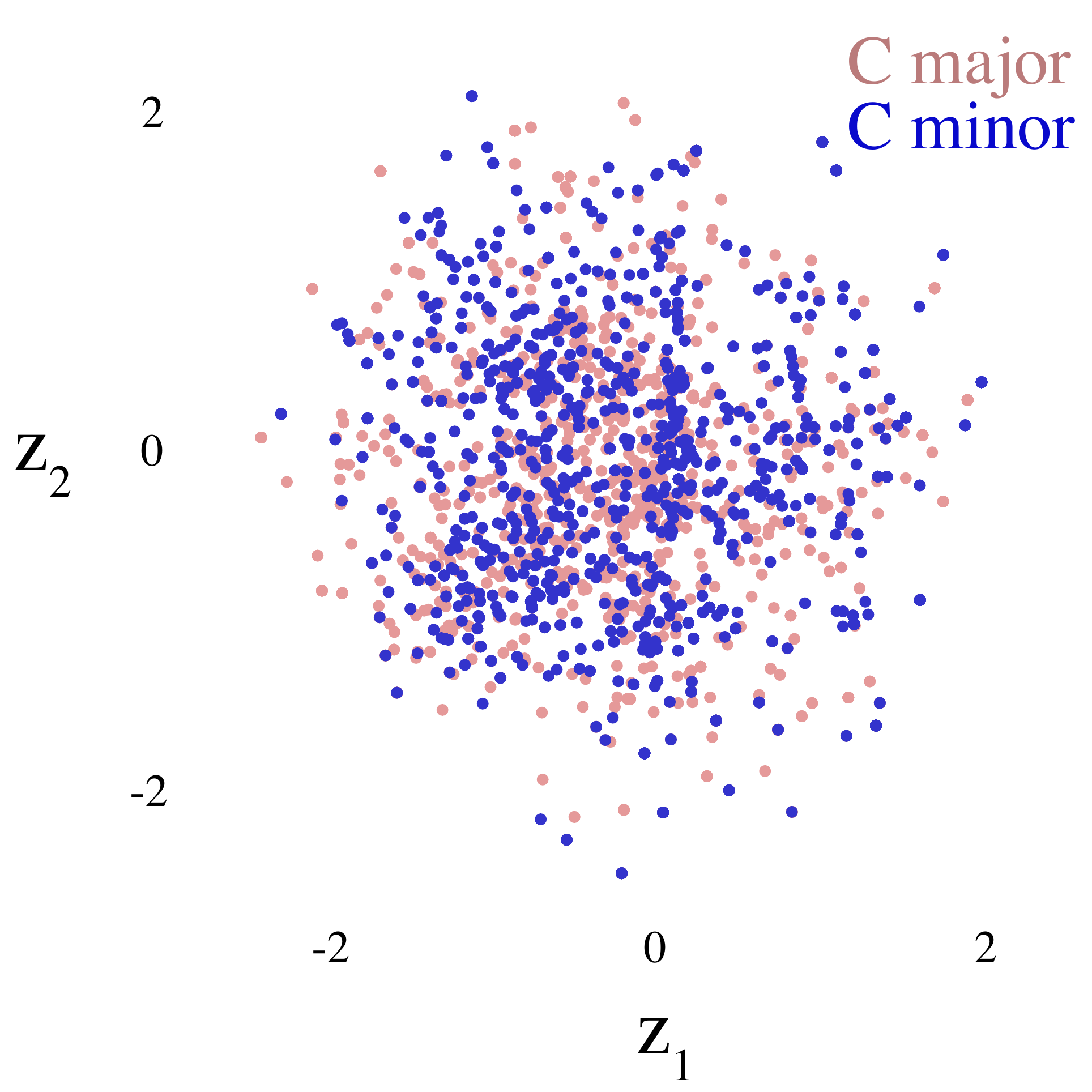}
\end{minipage}
 \caption{Comparison of the encodings, $\mu_z$, of held-out data for the VAE (left) and Classifying VAE (right) using inputs from songs in the key of either C major (unsaturated red points) or C minor (saturated blue points). The Classifying VAE encodes $z$ after first inferring the probability that $X$ came from a song in one of the two keys. The Classifying VAE's encodings more closely match the imposed prior $z \sim N(0, I)$ than do those of the VAE.}
 \label{fig:latentSpace}
\end{figure}

To prevent overfitting, training ceased when the loss on validation data did not decrease for five consecutive epochs. For each model and data set combination we performed a random hyperparameter search with $128$ runs \cite{bergstra2012random}, and identified the best hyperparameters in terms of performance on a held-out validation set. Performance was measured as the estimated average log-likelihood using the importance sampler described in \cite{rezende2014}.

% We performed a hyperparameter search using grid search, choosing the hyperparameters that resulted in the lowest loss on validation data (see Tables \ref{hyperTableJSB2}, \ref{hyperTableJSB10}, and \ref{hyperTablePiano}). For the VAE and Classifying VAE, batch sizes in the range of 10-50 were preferred, along with a latent dimensionality around 4-8. For the VAE+LSTM and Classifying VAE+LSTM, a batch size of 50-200 was sufficient, along with a latent dimensionality of around 8-16, and a sequence length of around 8-16. For all hidden perceptron layers (i.e., in the encoders, decoders, and classifiers) we found a dimensionality equal to 88, the input dimensionality, worked best. In both the VAE and VAE+LSTM models, we observed that the Classifying versions generally needed fewer latent dimensions than the standard versions. We attribute this to the fact that the extra Classifying network in the Classifying models adds predictive power and model complexity.

\textit{Sample generation.} To assess the key consistency of each model's generated samples, we provide each model with seed sequences $X^{seed}$ of length $T$, and use each model to generate the next $T$ timesteps. To generate music samples, the VAE and Classifying VAE use the last timestep of the seed sequence as $X_0$, and generate $X_t$ by sequentially autoencoding $X_{t-1}$ for $t = 1, ..., T$. The VAE+LSTM and Classifying VAE+LSTM generate samples similarly, except they first use the entire seed sequence to initialize the hidden states of the encoder and decoder LSTMs by passing the seed sequence through the network. Note that before generating music samples, the Classifying models first sample $w \sim q_\phi(w \mid X^{seed})$ from the classifier network and use this value of $w$ in the steps above.

% To generate a sample of length $T$ using a VAE, let $X_0 = X^{seed}$. Then for each $t \in \{1, ..., T\}$ we sample $z_{t-1} \sim q_\phi(z \mid X_{t-1})$ from the encoder, pass $z_{t-1}$ through the decoder to get $\pi_t = p_\theta(X_t \mid z_{t-1}, X_{t-2})$, and then sample $X^i_t \sim \mbox{Bernoulli}(\pi^i_t)$ for $i = 1, ..., 88$. To generate samples using the standard VAE+LSTM, we assume we are given a seed sequence sequence $X^{seed}$ also with length $T$, which we use to seed the hidden states of the encoder and decoder LSTMs. After doing this, let $X_0 = X^{seed}_T$. Then, for $t = 1, ..., T$, we iteratively sample $z_{t-1} \sim q_\phi(z \mid X_{t-1})$ and $\pi_t = p_\theta(X_t \mid z_{t-1}, X_{t-2})$ with $X^i_t \sim \mbox{Bernoulli}(\pi^i_t)$ for $i = 1, ..., 88$. Samples are generated for the Classifying VAE and Classifying VAE+LSTM as above, except that each model first samples $w \sim q_\phi(w \mid X^{seed})$ from the classifier, and provides $w$ to the encoder and decoder steps listed above.

\section{Results}
\subsection{Visualization of learned manifolds}

We start with a simplified example, comparable to previous experiments using neural networks to generate polyphonic music, in which the training data is the JSB Chorales corpus, with its songs transposed to be in either the key of C major or C minor \cite{rnn-rbm,STORN,vohra2015modeling,lyu2015modelling,johnson2017generating}.

\begin{figure}[ht]
\centering
\begin{minipage}{0.45\textwidth}
 \centering
 \textbf{$p_\theta(X \mid z=[-1,1], w=0)$}\\
\includegraphics[width=0.7\textwidth]{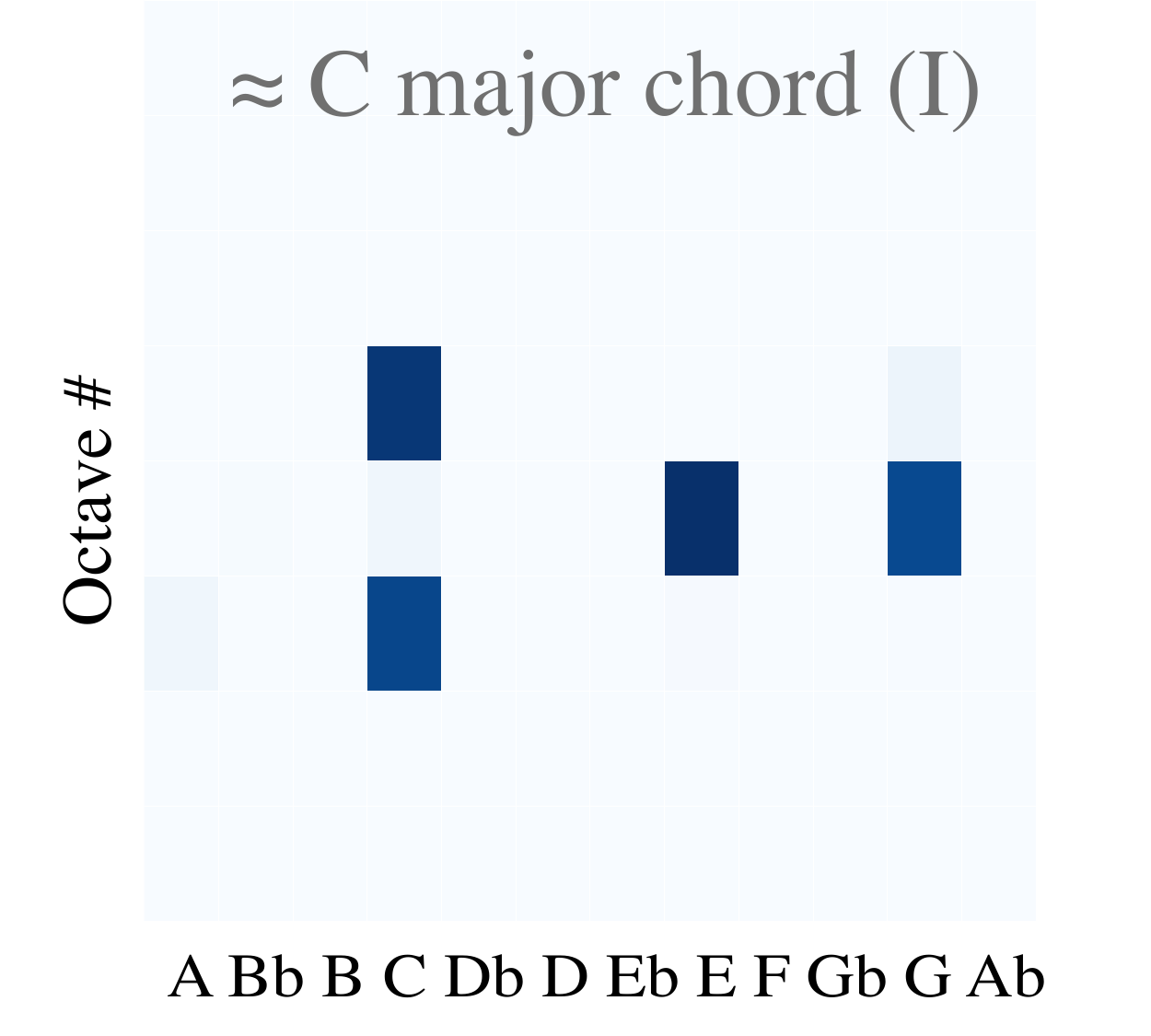}
\end{minipage}
\begin{minipage}{0.45\textwidth}
 \centering
 \textbf{$p_\theta(X \mid z=[0,1], w=0)$}\\
\includegraphics[width=0.7\textwidth]{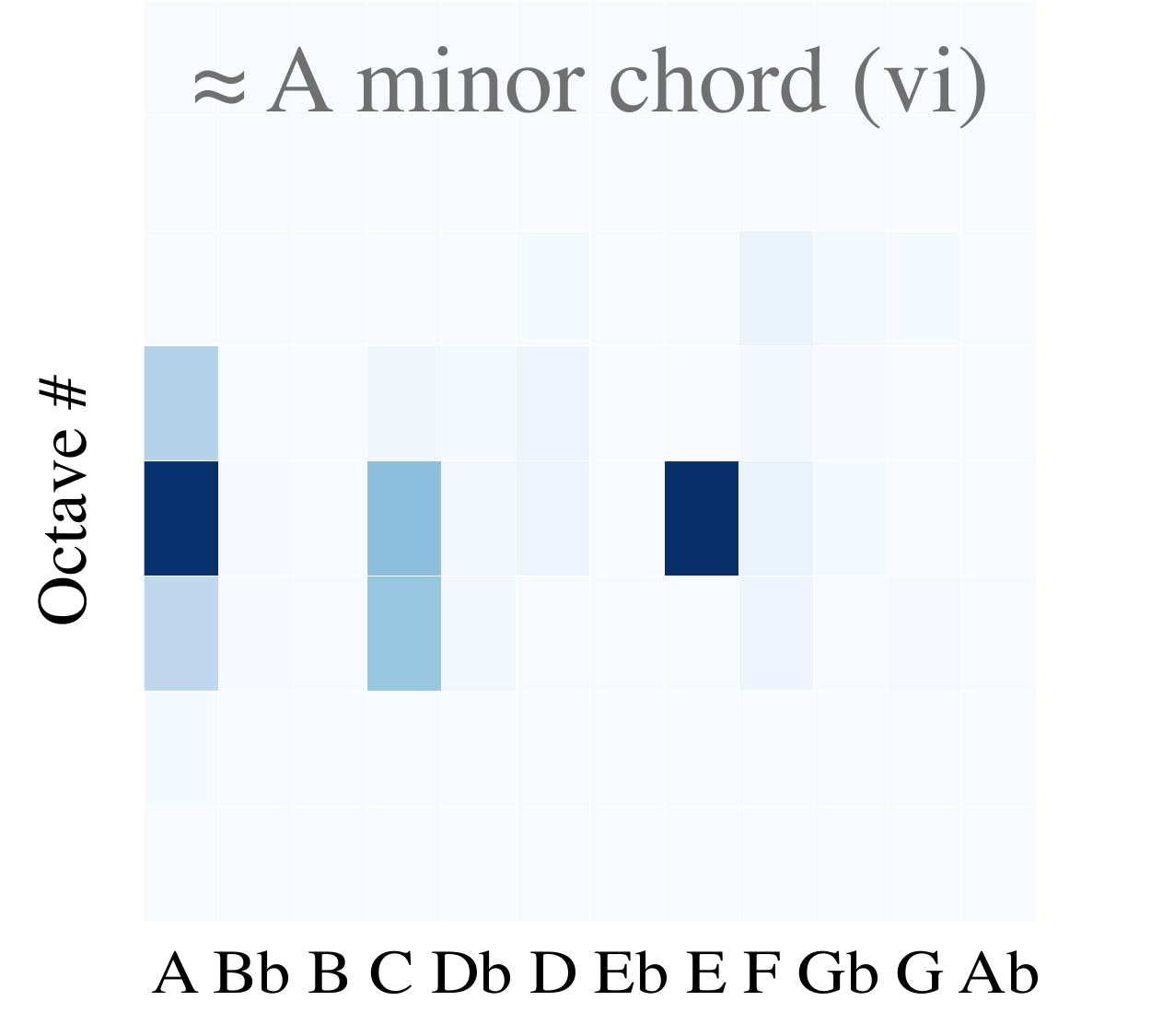}
\end{minipage}\\\vspace{5mm}
\begin{minipage}{0.45\textwidth}
 \centering
 \textbf{$p_\theta(X \mid z=[-1,1], w=1)$}\\
\includegraphics[width=0.7\textwidth]{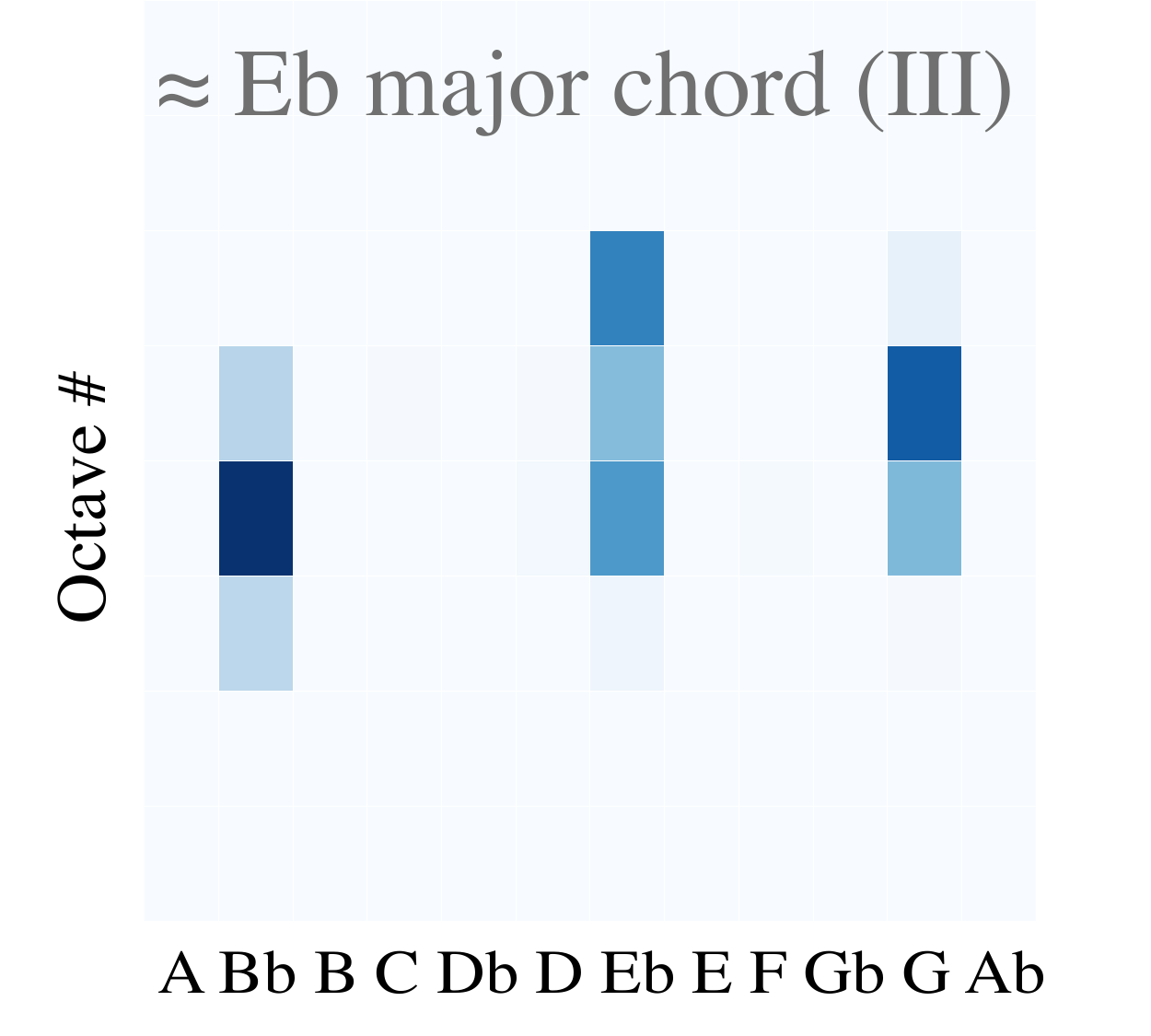}
\end{minipage}
\begin{minipage}{0.45\textwidth}
 \centering
 \textbf{$p_\theta(X \mid z=[0,1], w=1)$}\\
\includegraphics[width=0.7\textwidth]{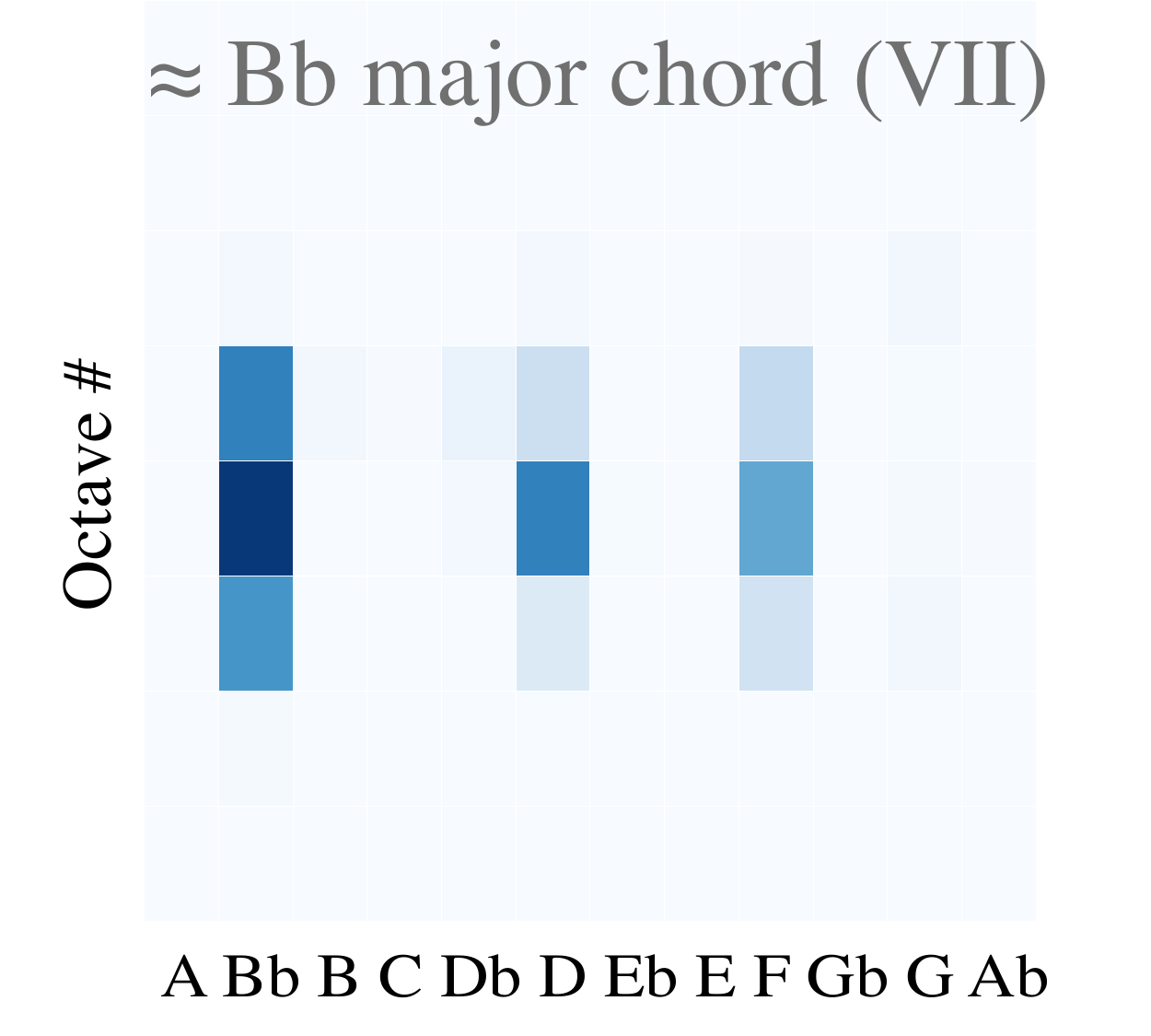}
\end{minipage}
\caption{Heatmap of decoded probabilities of $X$ given different values of $z$ and $w$, for the Classifying VAE in Figure \ref{fig:latentSpace}. In each panel, probabilities are arranged in terms of the corresponding pitch (x-axis) and octave (y-axis), with darker colors representing higher probabilities of playing the corresponding note. The text in each panel describes the most likely chord given the decoded probabilities. Overall, the value of $w$ controlled whether the decoded notes were appropriate for either C major (when $w = 0$) or C minor ($w = 1$), while varying $z$ and holding $w$ fixed resulted in different chords still consistent with the same key (e.g., the two panels in the top row).}
\label{fig:clvaeManifold}
\end{figure}

We trained a standard VAE and Classifying VAE where the dimensionality of the continuous latents, $z$, was two. Additionally, we modified the decoding model so that decoding at each time step is independent (i.e., the decoding model is not conditioned on $X_{t-1}$). This allows us to easily visualize the encodings learned by each network given a time step from a song in C major or C minor.

We visualize the encodings in both networks by depicting the mean encodings, $\mu_z$, of held-out data (Figure \ref{fig:latentSpace}). To understand how inputs in different keys were encoded, we color the mean encodings according to the key of each input (C major: unsaturated red; C minor: saturated blue). This reveals a critical difference in the encodings of the VAE and Classifying VAE: The encodings of the VAE show clustering based on the key of the input, while the Classifying VAE appears to more evenly utilize the latent space for songs in both C major and C minor. For example, for the VAE encodings in Figure \ref{fig:latentSpace} (left panel), values near the center (e.g. (0,0)) are more likely to be inputs from songs in C minor, while those near the edge (e.g. (0,2)) are more likely to come from songs in C major. This clustering is problematic when generating data, because we have no principled way of sampling from one cluster and not another. Over time, we will likely sample values of $z_t$ from different clusters, resulting in samples that alternate between C major and C minor. By contrast, the encodings of the Classifying VAE more closely resemble the prior distribution, $z \sim N(0, I)$, for songs in both C major and C minor. This is because the recognition model for $z$ depends on $w$, the current key. The absence of clustering in the Classifying VAE's encoding space suggests that the Classifying VAE may have better captured the data manifold than the standard VAE \cite{advAuto}.

One proposed benefit of modeling the key of the music is that we can control the key of the generated samples. We next assess whether this is the case with the Classifying VAE, by exploring whether the value of $w$ effectively controls the key of the generated samples. To do this, we visualize the outputs of the decoder network given different values of $w$ and $z$. The outputs, $p_\theta(X \mid z, w)$, represent the probabilities of different notes being generated, given the latent encodings. We visualize these probabilities as heatmaps, arranged so that two heatmaps in the same column depict the decoded probabilities given the same value of $z$ but different values of $w$ (Figure \ref{fig:clvaeManifold}). We identified the chord described by each heatmap, and found that the same value of $z$ resulted in vastly different chords depending on whether $w$ specified a song in C major (top) or C minor (bottom). This suggests that in the Classifying VAE, the value of $w$ controls the key of the generated notes, while different values of $z$ generate various chords consistent with that key.

\subsection{Log-likelihoods and key consistency}

We next show that the classifying networks in the Classifying VAE and Classifying VAE+LSTM enable these models to generate music that stays more in key than the standard models, while still maintaining similar average log-likelihoods. To assess this we trained a VAE, Classifying VAE, VAE+LSTM, and Classifying VAE+LSTM on four different data sets. This included (1) JSB Chorales with songs transposed to two keys (C major or C minor), (2) JSB Chorales with songs in their original keys, (3) Piano-midi.de with songs transposed to two keys (C major or C minor), and (4) Piano-midi.de with songs in their original keys. Here we did not restrict the models to using only two latent dimensions. For the Piano-midi.de data in the original keys, we discretized the music to sixteenth notes, while the data in two keys was discretized to eighth notes, for comparison to previous work. Because we were limited in computational power, for the Piano-midi.de data in the original keys we fit only the VAE and Classifying VAE models.

\begin{table*}[ht]
\centering
% \begin{tabular}{lllll}
% \hline
% \textbf{Data set} & VAE & Classifying VAE & VAE+LSTM & Classifying VAE+LSTM\\ \hline
% JSB Chorales (C major/minor) & -6.87 & -6.83 & -6.66 & -6.73 \\ \hline
% JSB Chorales (all keys) 	& -8.83 & -8.79 & -8.26 & -9.28 \\ \hline
% Piano-midi.de (C major/minor) & -7.51 & -7.49 & -7.05 & ? \\ \hline
% Piano-midi.de (all keys)   & -6.30 & -6.73 & ? & ? \\ \hline
% \end{tabular}
\begin{tabular}{lllll}
\hline
\textbf{Average log-likelihoods} & JSB Chorales & JSB Chorales & Piano-midi.de & Piano-midi.de \\
& (two keys) & (all keys) & (two keys) & (all keys)\\ \hline
VAE & -6.87 & -8.83 & -7.51 & -6.30 \\ \hline
VAE+LSTM & -6.66 & -8.26 & -7.05 & - \\ \hline
Classifying VAE & -6.83 & -8.79 & -7.49 & -6.73 \\ \hline
Classifying VAE+LSTM & -6.73 & -8.73 & -7.11 & - \\ \hline
\end{tabular}
\caption{Estimated average log-likelihoods on test data, for models trained on songs transposed to C major and C minor (``two keys''), or in the original, untransposed keys (``all keys''). All log-likelihoods were estimated using the importance sampler described in \cite{rezende2014} on held-out test data, after choosing the best hyperparameters using validation data.}
\label{modelComp}
\end{table*}

Table \ref{modelComp} displays our estimates of each model's average log-likelihood, which we computed using the importance sampler as described in \cite{rezende2014} on held-out test data, after choosing the best hyperparameters for each model using a validation set. Estimating the log-likelihood involves marginalizing out the unobserved variable $z$ (and $w$, for the Classifying models) in order to estimate $\log p_\theta(X)$, for all $X$ in the test data. Previous work has reported the performance of similar models on these corpuses after transposing the data to two keys \cite{rnn-rbm,STORN,vohra2015modeling,lyu2015modelling,johnson2017generating}. To our knowledge, the performance of our VAE+LSTM model on the Piano-midi.de data set (in two keys) is the highest reported for this data set, tied with the RNN-NADE model \cite{rnn-rbm}.

Somewhat surprisingly, we observed that both the VAE and Classifying VAE models achieved similar log-likelihoods as the VAE+LSTM and Classifying VAE+LSTM, for all data sets (Table \ref{modelComp}). This suggests that a model such as the VAE, which can use $X_{t-1}$ to decode $X_{t}$, is capturing almost as much of the temporal structure in these data sets as a VAE that uses LSTMs in the encoder and decoder networks. Given that music almost certainly has longer term dependencies, future models should aim to better capture these longer-term dependences.

Critical to our experiments, the addition of the Classifying networks to the standard VAE and VAE+LSTM resulted in similar performance in terms of log-likelihood, for both data sets. This was true even when fitting these models to data in the original keys. However, a model having high log-likelihood does not necessarily imply that its generated samples will be high quality \cite{theis2015note}. As we will show next, the generated samples of the Classifying models achieve higher key consistency.

\begin{table*}[ht]
\centering
\begin{tabular}{lllll}
\hline
\textbf{Key consistency} & JSB Chorales & JSB Chorales & Piano-midi.de & Piano-midi.de \\
& (two keys) & (all keys) & (two keys) & (all keys)\\ \hline
VAE & 86.7 $\pm$ 0.2\% & 74.3 $\pm$ 0.2\% & 76.4 $\pm$ 0.1\% & 60.1 $\pm$ 0.1 \\ \hline%& -7.51 & -6.30 \\ \hline
VAE+LSTM & 91.3 $\pm$ 0.2\% & 84.7 $\pm$ 0.2\% & 79.0 $\pm$ 0.1\% & - \\ \hline%& -7.05 & - \\ \hline
Classifying VAE & 90.1 $\pm$ 0.2\% & 85.9 $\pm$ 0.2\% & 76.9 $\pm$ 0.1\% & 67.0 $\pm$ 0.1 \\ \hline%& -7.49 & -6.73 \\ \hline
Classifying VAE+LSTM & 89.6 $\pm$ 0.2\% & 93.4 $\pm$ 0.1\% & 80.1 $\pm$ 0.1\% & - \\ \hline%& - & - \\ \hline
Classifying VAE* & 96.2 $\pm$ 0.1\% & 97.3 $\pm$ 0.1\% & 92.5 $\pm$ 0.1\% & 98.9 $\pm$ 0.0 \\ \hline%& -7.49 & -6.73 \\ \hline
Classifying VAE+LSTM* & 94.2 $\pm$ 0.1\% & 96.0 $\pm$ 0.1\% & 83.1 $\pm$ 0.1\% & - \\ \hline\hline
Data & 94.2 $\pm$ 0.1\% & 93.2 $\pm$ 0.1\% & 82.6 $\pm$ 0.1\% & 82.0 $\pm$ 0.1 \\ \hline%& - & - \\ \hline
\end{tabular}
\caption{Percentage of notes in generated samples consistent with a particular key (``key consistency'') (mean $\pm$ SE). Averages were computed as the geometric mean across samples. Each model generates a sample for $T = 16$ time steps after being seeded with each musical sequence in held-out test data. Chance performance is 67\%. The seed sequences from test data (``Data'') show a key consistency below 100\% due to variations in the key of each sequence. When the Classifying models are provided with the true key of the seed sample (models with $^*$) rather than using the inferred key (models without $^*$), this results in even higher key consistency.}
 \label{fig:inkey}
\end{table*}

We now assess whether the Classifying models are able to effectively improve the ability of the VAE and VAE+LSTM to produce music that stays in key. For each model, we generate music samples with length $T = 16$ after being seeded with each musical sequence of length $T = 16$ in the held-out test data of the JSB Chorales corpus. Although we use $T = 16$ here, other sequence lengths yield similar results. We then determine how consistent the generated samples are with the key of the seed sequence by calculating the proportion of notes in each sequence that are in the key of the seed sequence (Table \ref{fig:inkey}). For example, if the seed sequence is in C major, we count the proportion of pitch classes in the generated sample that are in the eight-element set $(A, B, C, D, E, F, G)$. Because each key is composed of eight pitch classes, and in total there are 12 pitch classes to choose from, chance-level performance is 67\%. However, even the seed samples from the test data contain notes not strictly in the eight-element set defining their key (Table \ref{fig:inkey}, ``Data''). For example, as suggested by Figure \ref{fig:pitches} (bottom panel), songs in JSB Chorales labeled as being in the key of C major also include a small proportion of the notes $Bb, Db, Eb, Gb$, and $Ab$.

We observe that, for the data sets in two keys, the key consistency of the Classifying models is comparable or better than that of the standard models (Table \ref{fig:inkey}, first column). However, for the data sets in the original keys, the Classifying VAE+LSTM model is able to generate musical samples with the same key consistency of the seed sequences, while the standard models generate much less consistent samples (Table \ref{fig:inkey}, second column). Finally, we note that providing our models with the true key of the seed sequence (rather than them having to infer the key) improves their performance above the other models in all data sets.

We also assessed two other statistics of the generated music for comparison to that of the held-out test data, for samples of length $T = 16$ from JSB Chorales. First, we measured the average number of notes played at each time step, which for the JSB Chorales test data was $3.9 \pm 0.0$ (mean $\pm$ SE). Second, we measured the tone span, or the average distance between the highest and lowest pitch in each sample, which for the JSB Chorales test data was $30.9 \pm 0.1$ (mean $\pm$ SE). For both of these metrics, all models achieved mean values within $6\%$ of that of the test data, on average. We provide these quantitative measures as a first approximation of the samples' musical quality, though we note that a more thorough assessment of musicality would involve subjective assessments by expert musicians \cite{collins2017computer}.

\section{Related Work}

There are a few other extensions to the variational autoencoder that allow for the data to be conditioned on a discrete class variable. This includes both the conditional variational autoencoder \cite{sohn2015learning,walker2016uncertain} and a semi-supervised variational autoencoder \cite{kingma2014semi}. Our work differs from these previous approaches as follows. First, in contrast to the conditional variational autoencoder \cite{sohn2015learning,walker2016uncertain}, we do not assume that we are provided with the conditioning class variable at test time (e.g., when generating samples). This would be beneficial in a live music generation setting. For example, our model could play along with a musician without having to be explicitly told which key to play in.

Though our model bears some similarity to the semi-supervised autoencoder \cite{kingma2014semi}, in fact the aims of these two models are quite distinct. The aim of the semi-supervised model is to improve classification accuracy using unlabeled data. By contrast, the primary aim of our model is to improve reconstruction error by leveraging a classifier trained on labeled data. Because we treat the class variable as a continuous value and use samples from its inferred posterior as inputs to the recognition network, our model has the option of using the classifier's outputs as an additional source of stochasticity during inference. This is because the aim of our model, unlike the semi-supervised autoencoder, is to optimize reconstruction error and not classification error. Also, though the semi-supervised variational autoencoder infers the class variables ($y$) for the subset of unlabeled data, the classifier and autoencoder layers must be trained separately. Because $y$ is multinomial, it must be marginalized out, and the generative likelihood (see their Equation 7) must be evaluated for each possible value of $y$ at each gradient step. As mentioned in their Discussion, this is a very expensive operation and means that training becomes more costly in the multi-class case. By contrast, we parameterize our latent class variable with a logistic normal distribution, which allows us to use the reparameterization trick, so that we can train both the classifier and autoencoder simultaneously.

\section{Conclusions}

We have developed extensions of the variational autoencoder (VAE) and recurrent variational autoencoder called the Classifying VAE and Classifying VAE+LSTM. This approach enables us to model the class of the data sequence using an additional classifier network. We demonstrate that this approach is effective for generating polyphonic music that stays in key better than do the samples from the standard VAE or VAE+LSTM models.

There are many interesting avenues for future work. In this paper we have made the simplifying assumption that the key of a particular musical sequence is fixed over the length of the sequence. Future work may attempt to predict the key at each time step and enable a method for predicting key changes. Another interesting future direction would be to classify a composer's style, where the class label $w$ would denote the composer's identity. As suggested by Figure \ref{fig:clvaeManifold}, changing $w$ while keeping the encodings $z$ the same might enable an effective method for translating a song in the style of Bach, for example, into a song in the style of Mozart.

% \bibliographystyle{plain}
% \bibliography{refs}
\end{document}